# Predictive process mining by network of classifiers and clusterers: the PEDF model

Amir Mohammad Esmaieeli Sikaroudi, Md Habibor Rahman

Abstract

In this research, a model is proposed to learn from event log and predict future events of a system. The proposed PEDF model learns based on events' sequences, durations, and extra features. The PEDF model is built by a network made of standard clusterers and classifiers, and it has high flexibility to update the model iteratively. The model requires to extract two sets of data from log files i.e., transition differences, and cumulative features. The model has one layer of memory which means that each transition is dependent on both the current event and the previous event. To evaluate the performance of the proposed model, it is compared to the Recurrent Neural Network and Sequential Prediction models, and it outperforms them. Since there is missing performance measure for event log prediction models, three measures are proposed.

Keywords: Process mining, Sequence prediction, Event log processing, Machine learning

## 1. Introduction

Process mining can be seen as gathering process intelligence, monitoring process activities, and analysis of processes employing event data. Process mining techniques have evolved significantly over the last decade. These techniques are helping organizations by providing data-driven information regarding the performance of the key processes, captured from different enterprise transaction systems. Process mining aims to rebuild a complete process model using available data from the real-time execution of processes (such as event logs, transaction logs) [1].

Process mining techniques, especially process discovery algorithms, can be used in the context of workflow management systems to model and describe an existing system, for example, patient records in hospitals, and transaction logs of enterprise resource planning systems. Using the event logs like patients' registration and billing in hospitals, process mining can identify homogeneous groups of patients that enable better comprehension of the process flow, prediction of workload, and creation of appropriate resource allocation plans [2]. Process mining can also support and automate increasingly complicated financial audits by modeling both data flow and control flow [3]. In the network domain, process mining techniques can comprehend the working relationships among multitudes of resources and identify key resources for process executions [4]. Moreover, process-oriented knowledge can be extracted from event logs of a manufacturing execution system that not only enables modeling the actual manufacturing process flow but also paves the way to conduct future experiments (such as machine analysis to find the utilization of machines) [5].

After extracting knowledge from a finite portion of historical data, process mining techniques can make use of the event log for generating predictive models. The created model helps to recommend possible future outcomes of a process while it is being executed and/or predict characteristics of process instances that are incomplete [6].

As a link between process-centric and data-driven approaches, process mining techniques can discover, ensure conformance, and enhance process models with/without using any a-priori information. In traditional process mining techniques, process models are generated by observing and analyzing the key activities/attributes of event logs that are directly related to the process. For example, a process model for the patient flow in a hospital can be created by considering patient flows within different departments

(activities) to form clusters of typical pathways of patients. The key activity which is patients' movement across different departments in the hospital will help us understand the overall patient flow, and later enable us to predict the movement of future patients. However, the performance of future predictions can be improved by incorporating additional attributes of patients such as their age, if available, in the model. The rationale here is that although age doesn't dictate the possible movement of a patient in a hospital, age can be an influencing factor for the services which a patient requires, and thus it may affect their movement across departments. Note that, such attributes are often ignored in existing process mining techniques. Therefore, we intend to investigate the prospect of incorporating such additional attributes within the process mining technique, for a model to learn better and make better-informed decisions in predicting future events.

In a nutshell, as a link between process-centric and data-driven approaches, process mining techniques can discover, ensure conformance, and enhance process models with/without using any a-priori information. In traditional process mining techniques, process models are generated by observing and analyzing the key activities/attributes of event logs that are directly related to the process. For example, a process model for the patient flow in a hospital can be created by considering patient flows within different departments 

## 2. Background

### 2.1. Markov chain

Markov chain (MC) refers to a stochastic model that can describe the potential sequence of future events by utilizing the information observed from the previous sequence of events. In Markov models, future events can be predicted from the transition probability that is governed only by the present state of the system and the amount of time elapsed in the system. Similarly, process mining techniques assume that certain information can be extracted from the available event log. Therefore, Markov models are often used for process mining. However, Markov models may not work efficiently with incomplete or irrelevant data present in the event log. Consequentially, sequence clustering techniques have also been applied in process mining, that uses first-order MCs, to weather noise in the data, providing more flexibility. Moreover, elaborate models such as Hidden Markov Models are also used for apprehending additional information about different stages in a process. HMMs model observed sequences as probabilistic functions of unobserved sequences. In contrast to Markov model-based approaches where sequences are often drawn from discrete state space, HMM-based models are scalable in time. HMM-based clustering techniques find the probability of traces being generated from each HMM and then try to find the most likely cluster based on the trace [7]. Yet, HMMs become less efficient with the set of possible hidden states growing large, often computationally infeasible to model long-range dependencies [8].

### 2.2. Petri nets

Petri nets are strong modeling language which utilizes tokens and conditional transitions. If multiple transitions can be activated, Petri net will be stochastic. Nondeterministic property of stochastic petri net

(SPN) has Markovian property and it can be expressed as a MC, however, Petri net provides more complex and in-depth model compared to MC. Petri nets are usually used to express the flow of a system to evaluate its performance. Lyu, Chen [9] used Petri net to evaluate a failure detection and recovery system designed for mobile agents which could check if the next destination server is alive by utilizing Petri net logics. Zhong and Qi [10] studied on transformation of business process execution language of web services into SPN. It is shown that by utilizing SPN reliability, workload, and execution time of a web service can be evaluated. Rogge-Solti and Weske [11] proposed a predictive model for remaining time to the next event. They simulated the SPN of the system for unknown portion of a partially known case. The average value of the durations is used to predict the remaining times.

### 2.3. Recurrent neural network

Recurrent neural networks (RNNs) are powerful learning models that can capture the dynamics of sequences through recurrent cycles within the "network of nodes" [8]. For modeling data with sequential structures, that have inputs and outputs of variables lengths, RNNs come handy with the capability of selectively passing necessary information across different steps sequentially during processing data elements one by one. As a result, RNNs can model both temporal and sequential data simultaneously. Moreover, compared to traditional neural networks, RNNs relax the vital assumptions of data independence as well as, the fixed input and output dimension. Most importantly, RNNs can overcome the key limitations of Markov models and capture "long-range time dependencies" [8]. In contrast to HMMs, hidden states at a particular time step are capable to hold information from a sufficiently large context window in RNNs.

### 2.4. Process mining

Process mining is a family of techniques that allow extraction of information from events and transaction logs, for developing models that can describe a process. Very similar to data mining, it refers to the utilization of data available in the information system of an organization for visualization and decision making [12]. Data stored in IT systems and event logs can be leveraged to develop models without any prior information (process discovery), compare an existing model with that process's event logs (conformance of a process), and improve any existing models using captured information from event logs (process enhancement) [13]. The output of process mining is usually expressed by a Petri net.

### 2.5. Sequential pattern mining algorithms

Mining sequential patterns from a large database of sequences result in more complex problems compared to discovering frequent itemset and association rule mining, due to the additional constraint of the ordering of items. Yet, this approach of finding out the set of all possible sequences in a database has been used widely in myriads of fields for analyzing web access patterns, consumer behavior, biological sequences, telecommunications, and so on [14] Sequential pattern mining algorithms can be broadly classified into two groups: a. algorithms based on apriori technique and b. algorithms based on the pattern growth technique. For instance, Generalized Sequential Patterns (GSP), Sequential Pattern Discovery using Equivalence classes (SPADE), and Constraint-based Apriori algorithm for Mining Long Sequences (CAMLS) algorithms are based on the apriori technique. On the other hand, examples of algorithms based on the pattern growth technique are Prefix-projected Sequential pattern mining (PrefixSpan), MEMory Indexing for Sequential Pattern mining (MEMISP), LAst Position Induction (LAPIN), and PRIme-Encoding Based Sequence Mining (PRISM). Febrer-Hernández and Hernández-Palancar [14] compared different sequential pattern mining algorithms.

### 2.6. SPMF

Sequential Pattern Mining Framework (SPMF) is a data mining library implemented in Java, which is specialized in frequent pattern mining, for realizing relationships and patterns in databases [15]. SPMF has been widely used in recommender systems, mining web usage, understanding learning behavior, anomaly detection, forecasting, and sequence predictions. SPMF offers myriads of algorithms (196 data mining algorithms) that are capable to work on transaction database and sequence database. From a data mining viewpoint, algorithms offered by SPMF can perform frequent itemset mining, association rule mining, sequential pattern mining, sequence prediction, time-series mining, clustering, and classification [16].

### 2.7. Related works

For forecasting the performance of unfinished process instances, Folino et al. (2012) presented an improvised Performance Prediction Model (PPM) with an additional predictive clustering approach. Based on context features, homogeneous execution clusters are recognized and then specific performance-prediction models were developed for each cluster [17]. Various sequential pattern mining and evaluation measures have been applied in process mining filed to predict future events. Cuzzocrea, Folino [18] reviewed different aspects of process mining such as discovery algorithms and prediction models. They proposed clustering to predict performance of a system and time-series to predict future values. Di Francescomarino, Dumas [19] proposed an interesting approach that encodes event sequences into a control flow. A clusterer groups similar flows together and a classifier predicts a partial sequence to determine which cluster of the control flows it should be assigned to. Technically different flows in the system are preprocessed and the clusterer simplifies the process by grouping similar flows and the classifier assigns a partial case into a group of flows. Prediction of events and durations has been studied by RNNs because of the flexibility of the algorithm to handle generic datasets and predict different features. Tax, Verenich [20] used long-term short-term neural networks to predict remaining time of a case and the next activities. They used a prefix to express the length of a case and reported mean absolute error for different length sizes. The RNN model is tested on business process intelligence challenge 2012. To avoid predicting overtly long cases, at the end of each case a symbol was added to represent the end of the case to be learnt by the model. Klinkmüller, van Beest [21] used rule-based classification model to predict next events based on generated synthetic data. They investigated the effect of different data characteristics such as case length. Teinemaa, Dumas [22] proposed a method to handle both structured and unstructured data in event log dataset. Their proposed method first applies text mining to label unstructured text into positive/negative. Then a classifier predicts the control flow of encoded sequence. For each possible length of cases a feature extracting model and classifier is made.

Discrete sequence prediction can play a vital role in information-theoretic applications, dynamic program optimization, adaptive human-machine interfaces, and anomaly detection systems. As an early work in predicting discrete sequences, Philip (1992) presented the transition directed acyclic graph (TDAG) learning algorithm [23]. TDAG learns by generating an initial tree consisting of the root node only. Based on new observations, successors of the initial node (knows as children) are generated in the form of a graph. The rarely visited nodes referring to infrequently occurring incidents are trimmed from the graph and the height of the graph is also bounded for preventing unbounded growth of the tree i.e., the state space. Predictive models have also been used to foresee the web surfing pattern of users by using the distribution of visits over various web pages. Pitkow et al. (1999) proposed the all-k-order Markov model (AKOM), which fits semi-Markov models of up to the order k on the training set, for envisioning the next web page by learning from the lengthiest repeating subsequences in the historical data i.e., previous web sessions [24]. The obtained distribution of visits helps re-weighting and re-ranking results in text-based search engines. Moreover, multiple tools have also been developed to recommend related pages to the user [25]. Kinnebrew et al. (2013) used action abstraction which is a sequence mining technique, to identify and compare students' learning behavior patterns [26]. Gueniche et al. (2013) proposed the Compact Prediction Tree (CPT) model for predicting sequential items over a finite state space of alphabets [27]. They reported significant improvement in accuracy and training time. CPT differs from Prediction by Partial Matching (PPM) and All-K-Order Markov (AKOM) in terms of efficient utilization of training sequence information.

In 2015, for resolving the time and space complexity of CPT, Gueniche et al. presented the CPT+ model which was around five times faster than CPT [28].

Researchers have also implemented the Ziv-Lempel algorithm of 1978 for data compression, with applications in different fields including linguistics [29, 30]. The LZ78 lossless data compression algorithm replaces recurrent occurrences of data with predefined coded references to a dictionary which is built using the input data stream. References contain starting positions as well as lengths of earlier occurrences. Amirat et al. (2017) modeled vehicle movements with a dependency graph, an approach that was primarily developed for efficient resource prefetching mechanisms on the Internet. Using graphs to characterize road networks, they generated a prediction graph depending on the current location and historical data of a vehicle. The prediction graph is capable to predict the potential route of a driver comparing the vehicle's current trajectory with graph paths [31].

## 3. Problem definition

At first, the format of the data should be clarified. The data format in process mining and event log processing is trivial. Therefore, in this paper extensive notations are avoided and the reader should refer to the literature, however, it is required to understand the basics to track the contribution in this paper. consists of three major attributes: case, event, and timestamp. If the timestamp attribute is available, the duration of the processes can be studied. However, the timestamp is not mandatory as far as a sequential pattern exists in the records. On the other hand, the case attribute is an important feature that distinguishes the data from time-series datasets. The case attribute also separates the sequences into small portions indicating independence among them. In other words, the data need to contain several short sequences or time-series, separated by the case attribute. Therefore, the data can be represented by several time-series or several sequences. The event attribute addresses the discrete events that occurred in the system. Events can be represented as states in a system, which consequently can be represented by state transition models. In addition to the major attributes, other relevant attributes of the data (such as age as a numeric attribute, gender as a nominal attribute) can provide extra information from the event logs. Despite that the additional attributes can be used for making more informed decisions and enhancing the performance of the predictions, a solid model to handle the prediction of all attributes together is missing in the literature.

In this research, the goal is to make a model for predicting events, durations, and extra features altogether, which is named as the PEDF model. Here, PEDF is the abbreviation of predict, event, duration, and features. The model has similarities to state transition models such as SPN and MC but each state and each transition is made of standard machine learning models. The model in contrast to the Markovian properties is not memoryless and it adds more depth to it. Furthermore, the proposed model can be built modularly which can provide significant flexibility and performance.

## 4. Methodology: the PEDF model

The proposed methodology (PEDF) is based on a network consisting of nodes representing the events (states) and links representing the transitions. Dataset is a log file that should be processed to extract events and links. To each case in the log file, two dummy events are added, i.e. "start" and "end".

Two main assumptions are governing the proposed model. First, some information is added to the entity in the system on each transition. Therefore, each link should maintain a distribution of the changes when transiting from a source node to a destination node. Links have directions, and between each pair of nodes, maximum two links can be established i.e. forward and backward. Second, each node should maintain a distribution about the links where the entities enter and leave a node. By manipulating the two sets of information, a predictive model can be constructed for iteratively predicting an entity in the system to the end.

### 4.1. Data manipulation

Residing duration and extra features in the nodes are cumulative to represent the memory of the case. Cumulative numeric features are straightforward, but for nominal features, the latest value is considered. For instance, consider that the value for a nominal feature is "A" on event 1. When the value doesn't change on event 2, it remains "A". But if there is a change in event 3, the value changes to "B" and it remains the same until it changes again in another upcoming event. Therefore, at each node, all features are not necessarily changed. For a customer in a shop, for example, the event "checking a product" does not add to the payment amount, but cumulative payment is the amount that the customer paid, starting from the time he/she entered the system, till the exit from the system.

In contrast to the nodes, data stored in the links are the differences between two events in the log file. Consider, for a specific case, the transition from event 1 to event 2 adds 10 days to the timestamp and 20 units to the payment feature and changes the value of a nominal feature from "A" to "B".

### 4.2. Model formulation

Each node in the PEDF model should adaptively decide on the next link and cluster, after observing the incoming link and cluster. Each link generates clusters to facilitate the possibility to iteratively generate future predictions. The centroid of the predicted cluster is added to the case, on each transition to the next nodes. The major notations of the proposed model are shown in Table 1.

Table 1, The notations of the research.

| Notation | Description |
|---|---|
| $P_{i,j}$ | Probability of transition from node $i$ to node $j$ |
| $N_i$ | Node $i$ related to event $i$ |
| $LC_{c,i,j}$ | Cluster $c$ for the link connecting node $i$ to node $j$ |
| $LCC_{c,i,j}$ | Cluster centroid $c$ for the link connecting node $i$ to node $j$ |
| $EF_i$ | Cumulative extra features in node $i$ |
| $DU_i$ | Cumulative duration in node $i$ |
| $RE_{s,t}$ | Real event $t$ in case $s$ |
| $PE_{s,t}$ | Predicted event at time $t$ in case $s$ |
| $RD_{s,t}$ | Real duration for event $t$ in case $s$ |
| $PD_{s,t}$ | Predicted duration for event $t$ in case $s$ |
| $REF_{s,t}$ | Real extra features for event $t$ in case $s$ |
| $PEF_{s,t}$ | Predicted extra features for event $t$ in case $s$ |
| $BPL_s$ | Binary value to indicate if the predicted case has more events than real case for case $s$ |
| $EE$ | Event error |
| $DE$ | Duration error |
| $FE$ | Error for extra features |

Each node should handle the probability of transitions. $P_{ij}$ is the transition probability from event $i$ to $j$, which is the probability of transition by the link connecting node $i$ to node $j$, through cluster $c$. The probability of the transition is conditional to the previous input link that connects node $k$ to node $i$ with cluster $c'$, current node $i$, cumulative durations, and cumulative extra features, as shown in the formula (1). Therefore, the model represents a one-layer memory state transition instead of the memoryless transition in MCs which also considers the extra features in cumulative format.

$$P_{i,j} = P(LC_{c,i,j}|N_i, LC_{c',k,i}, DU, EF) \qquad (1)$$

After selecting the most probable $P_{i,j}$, based on formulas (2) and (3), $EF_j$ and $DU_j$ are updated.

$$EF_j = EF_i + LCC_{c,i,j} \qquad (2)$$

$$DU_j = DU_i + LCC_{c,i,j} \qquad (3)$$

The steps of PEDF can be summarized in the following steps:

*Initialize*
- Generate the network based on training log files
- Assume a portion of the case is known
1. The next most probable outward link and cluster is selected based on cumulative features
2. The cluster's centroid is added to the cumulative features
3. The entity traverses to the next node
4. **if** the next node is "end"
5.     the case prediction has finished
6. **else** jump to step 1

*end*

The PEDF model not only can generate predictions for future events but also can predict timestamps and extra features. Additionally, parallel processing can be applied to the proposed model. Note that, the training phase of the suggested model is accomplished in two steps. First, all links need to be clustered using the difference in their corresponding data. Second, all nodes are trained by their corresponding cumulative data. Each link is clustered independently, and each node can be trained independently from other nodes as long as its dependent links are clustered. The PEDF approach is made by standard clusterers and classifiers from the Weka machine learning package [32]. The schematic view of the PEDF model can be seen in Figure 1.

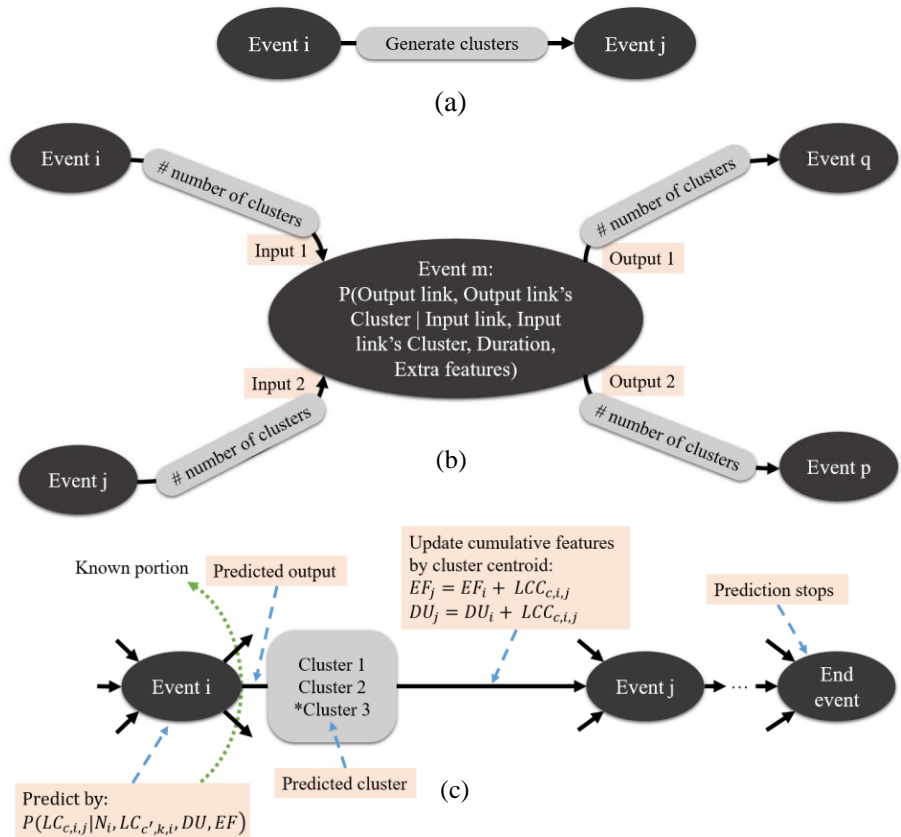

Figure 1, Schematic view of the PEDF model's mechanism. (a) Clusterer generates clusters for each link. (b) Classifier generates the predictive model to decide on the next events. (c) Prediction cycle of a partially known case.

### 4.3. Comparison of the existing and the proposed approach

In this section, at first, the similarities and differences of PEDF to other approaches are discussed. Then two approaches are discussed to compare with the proposed algorithm.

#### 4.3.1. Difference between time-series and process mining

Since timestamp is available in this research, it has similarities to time-series analysis. Time-series data refers to sequential data points captured at a certain time, usually at a fixed time interval. The high volume of automatically generated time-series data (such as operating cost, traffic, income) can be used for anomaly detection, trend analysis, association analysis, forecasting, and predictive analysis through time-series data mining techniques. Time series data mining focuses on the analysis of large static datasets or big data to transform embedded connections and/or relationships in those datasets for extracting valuable information. Traditional data mining procedures (such as clustering, association rule learning, regression) are not concerned about the business process model, rather focus on analyzing certain steps of the overall process [13]. Both data mining and process mining techniques are propelled by an urge to reveal hidden patterns of available data, for gathering business intelligence and aid the decision-making process. However, process mining is distinguishable in the sense that it emphasizes on realizing, monitoring, and refining actual business processes, while data mining has no direct connection to business processes. Furthermore, process mining is different from time-series because it contains several short independent time-series but not a bulky series. Besides, process mining search patterns for causal processes as well as, look for answers to specific questions. Contrarily, data mining is limited to pattern recognition but does not answer particular questions regarding causal processes. Moreover, data mining techniques often exclude data showing deviation from the usual patterns which are included in process mining as an opportunity for future improvement. In a nutshell, process mining incorporates data mining and process modeling, yielding to live models connected to the business that can be updated as required.

#### 4.3.2. Difference between SPN/MC and PEDF

Since the proposed method has states and transitions, it is similar to state transitioning models such as MCs. However, there are extra features that affect the transition probabilities. The dataset can be represented as a MC with one strong transient state called "Start" and one absorbing state called "End" to show the start and end of a sequence. The Markovian assumption does not hold for the proposed method and the transition matrix is more complex as well as, each state transition affects the next transition. Therefore, the proposed model is not memoryless.

Usually, Markovian models introduce more states when considering nominal features such as gender, i.e. a new set of states is generated for males and another set for the females. Similarly, for a numeric feature, first, it is binned into categories and the same state generation procedure is applied. However, the shortcoming of this procedure is that the incoming data affect the state space of the model which makes the model large. Coming back to the hospital example in the introduction, one state can be represented by "Move patient to ward/heart rate 90-110/blood pressure 120-125/…". Exhaustive combinations of the states will lead to a very large model. Another disadvantage of NPN/MC lies in simple and memoryless transition probabilities which is prone to overfitting for prediction of future states.

#### 4.3.3. Recurrent neural network

In order to use RNN, first, the data should be converted into a three-dimension matrix consisting of event sequences, features, and case dimensions. RNN has several available implementations. Deeplearning4J is the commonly used library for testing RNN. Each attribute is normalized for training and denormalized after testing the model. The data format is cumulative which means that the neural network predicts the cumulative features on each predicted event.

RNN model uses all features as both input and output. Assuming a portion of a case is known, each time RNN predicts the next event to be added to the model, and the next prediction happens. This process stops unless the "end" event is met or more than 10 events are predicted. The RNN model uses 3 long short-term memory (LSTM) layers with 200 neurons on each layer. Squared loss, sigmoid activation function, and maximum iteration of 100 are used to build the model.

Since each case has a different length of sequence, all sequences are padded and masked. Data is padded to the longest sequence with extra zeros. However, there could be zero values in the actual dataset that can potentially be mistaken by padded values. Therefore, masking is required to distinguish empty inputs/outputs in the dataset.

### 4.3.4. Sequence prediction

Sequence prediction models don't use timestamps and other extra features. The algorithms CPT+, PPM, DG, AKOM, and LZ78 in SPMF collection are used. The log data needs to be converted into a proper format first. Assuming that a portion of a case is known, the model predicts the next sequence. The prediction is added to sequence for next predictions until the "end" event is reached or more than 10 events are predicted. Sequence prediction models are unable to use and predict the duration as well as extra features.

## 5. Results

To assess the effectiveness of the PEDF model, real-life event log dataset "Road Traffic Fine Management Process" is used [33]. To reduce the runtime, only 50,000 records are selected from the beginning of the file containing 14333 cases and the longest case has 12 events and the average length of the cases is 4.52. The dataset can be seen in Figure 2 which shows a log file as an example of the process mining application. The cases are randomly partitioned into 70% training and 30% testing.

Figure 2, "Road Traffic Fine Management Process" dataset.

To evaluate the proposed model, standard and well-known algorithms are used. PEDF model's classifiers are Naïve Bayes (NB) [34], Random Forest (RF) [35], and Multi-Layer Perceptron (MLP) [36]. The clusterers are Canopy (CA) [37], Expected Maximization (EM) [38], K-means (KM) [39], and Cascade K-means (CKM) [40].

To compare the algorithms, three measures are defined i.e. event error, duration error, and feature error. The measures can be summarized in formulas (4), (5), (6), and (7). Formula (4) is based on the idea that as long as the real event sequence and the predicted event sequence have the same length, it is checked if they are the same. If the real sequence or the predicted sequence is longer, for each extra event, error is added by one. A similar concept is applied to duration and features on formulas (6) and (7) but it requires distinguishing which sequence is longer, the real sequence, or the predicted sequence. The reason is that the feature error is based on the feature values, but the event error is based on event nonconformity. Therefore, the binary variable $BPL_s$ in formula (5) is used to get the longer sequence.

$$EE = \sum_s \left[ \left( \sum_{t=1}^{\min\{t \in RE_{s,t}, t' \in PE_{s,t'}\}} |RE_{s,t} - PE_{s,t}| \right) + \max\{t \in RE_{s,t}, t' \in PE_{s,t'}\} - \min\{t \in RE_{s,t}, t' \in PE_{s,t'}\} \right] \quad (4)$$

$$BPL_s = \begin{cases} 0 & if\ t \in RE_{s,t} > t' \in PE_{s,t'} \\ 1 & if\ t \in RE_{s,t} < t' \in PE_{s,t'} \end{cases} \quad (5)$$

$$DE = \sum_s \left[ \left( \sum_{t=1}^{\min\{t \in RE_{s,t}, t' \in PE_{s,t'}\}} |RD_{s,t} - PD_{s,t}| \right) + \sum_{t=\min\{t \in RE_{s,t}, t' \in PE_{s,t'}\}+1}^{\max\{t \in RE_{s,t}, t' \in PE_{s,t'}\}} RD_{s,t} * (1 - BPL_s) + PD_{s,t} * BPL_s \right] \quad (6)$$

$$FE = \sum_s \left[ \left( \sum_{t=1}^{\min\{t \in RE_{s,t}, t' \in PE_{s,t'}\}} |RF_{s,t} - PF_{s,t}| \right) + \sum_{t=\min\{t \in RE_{s,t}, t' \in PE_{s,t'}\}+1}^{\max\{t \in RE_{s,t}, t' \in PE_{s,t'}\}} RF_{s,t} * (1 - BPL_s) + PF_{s,t} * BPL_s \right] \quad (7)$$

Table 2, Comparing SPMF and RNN algorithms.

| Known % | RNN | | | CPT+ | DG | PPM | AKOM | LZ78 |
|---|---|---|---|---|---|---|---|---|
| | Event error | Duration error | Feature error | Event error | | | | |
| 20 | 19,056 | 1,935,005.7 | 1,908,399.0 | 15,168 | 12,664 | 11,682 | 11,682 | 11,682 |
| 30 | 17,245 | 1,756,719.0 | 1,807,048.6 | 15,143 | 12,664 | 11,682 | 11,682 | 11,682 |
| 40 | 11,978 | 1,451,381.4 | 1,663,294.3 | 14,873 | 12,655 | 11,673 | 11,671 | 11,673 |
| 50 | 10,883 | 1,323,388.9 | 1,592,824.4 | 12,804 | 12,605 | 11,605 | 11,590 | 11,597 |
| 60 | 10,630 | 1,300,882.4 | 1,477,401.9 | 10,960 | 12,572 | 11,579 | 11,541 | 11,565 |
| 70 | 5,059 | 978,792.6 | 1,324,451.8 | 10,687 | 12,476 | 11,375 | 11,283 | 11,285 |
| 80 | 3,548 | 424,237.2 | 1,191,841.1 | 4,157 | 6,644 | 4,220 | 4,119 | 4,119 |
| 90 | 2,537 | 379,796.5 | 903,359.4 | 2,468 | 2,584 | 3,410 | 3,412 | 3,410 |

Based on proposed measures, the performance of SPMF and RNN algorithms are shown in Table 2. The performance of the PEDF model is shown in Table 4. Based on the results, PEDF shows significantly better performance. RF with KMeans under fixed 50 clusters shows the best performance among various tests on classifiers and clusterers. Figure 3 shows the compared algorithms when half of the cases' length is known. RNN shows better performance for duration especially when a large portion of a case is known. On the other hand, RNN fails to predict extra features. Canopy clusterer shows significantly poor performance because it generates very few clusters in each link. Naïve Bayes shows poor performance on all measures specially along with Canopy clusterer. Random Forest and Multilayer Perceptron generally perform better when little information is known about cases and they exhibit more stable performance. SPMF models don't show significant improvement until 80% of case lengths are known.

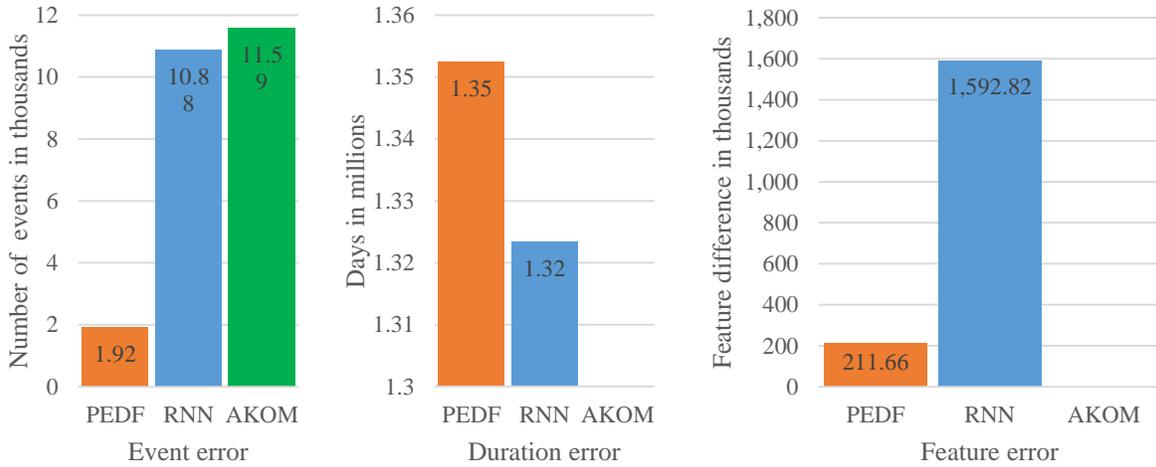

Figure 3, Comparison of PEDF with other algorithms by 50% known case length.

Training runtime for PEDF, RNN, and SPMF models are compared in Table 4. PEDF and RNN used 8 processing threads of AMD Ryzen 7 1700X processor and SPMF models were run by single thread. Data reading, preprocessing, and sampling are not considered in training runtime.

Table 3, Runtime comparison of PEDF, RNN, and SPMF models in seconds.

|  | PEDF | | | | RNN | SPMF | | | | |
| --- | --- | --- | --- | --- | --- | --- | --- | --- | --- | --- |
|  | Expected Maximization | Canopy | CascadeKMeans max 100 clusters | Kmeans 50 clusters |  | CPT+ | DG | PPM | AKOM | LZ78 |
| NB | 480 | 2 | 442 | 3 | 1408 | 0.08 | 0.018 | 0.018 | 0.053 | 0.088 |
| RF | 495 | 5 | 458 | 11 |  |  |  |  |  |  |
| MLP | 560 | 30 | 2369 | 1255 |  |  |  |  |  |  |

## 6. Conclusion

In this research, a new algorithm is developed to process event logs, convert them to a network, and predict partially known cases in event log datasets. The proposed method not only predicts future sequences but also use and predict extra features to enhance the performance of the prediction. The PEDF model is based on the assumption that state transition is dependent on the current state, previous state, clusters, and extra features. Therefore, PEDF is a one-layer memory-based state transition model. One-layer of memory and utilizing machine learning classifiers help the model to overcome overfitting while memory-less Markovian models apply simple transition probabilities which prone next state transition to overfitting. In this model, the log dataset is processed into two data types, i.e. difference data which is stored in the links, and cumulative data which is stored in the nodes. The source code of PEDF and compared models can be found on GitHub repository [41]. PEDF model fills the gap between time-series, sequence mining, and event processing models and in contrast to deep neural networks, PEDF model is made of generic classifiers and clusterers. There are well-established methods to assess performance for a typical machine learning problem such as average prediction on cross-validation, but for event log datasets new approaches are required. In this paper the percentage of partially known cases and different error types were used to assess model performance. Future researches can investigate more reliable measures to benchmark models.

One of the limitations of the proposed method is that a cluster centroid is added to partial known cases on each transition. If the number of clusters is low, significant information will be lost and low-quality information is generated iteratively which may affect next event predictions. Contrarily, if the number of clusters is high, the classification task will become increasingly difficult and can lead to poor accuracy. Therefore, there is a trade-off in the model, regarding the selection of the number of clusters, which needs to be investigated further. However, generally, with higher number of clusters the model performs better compared to low number of clusters. Furthermore, instead of cluster centroid, other measures can be used such as a random value around the dense regions of the cluster.

The strength of the PEDF algorithm lies in its modularity. This means that not only each section of the model can be trained in parallel but also partial data of a case can be fed into only one node and its connecting links to update only a portion of the network using new incoming data. Furthermore, if on future log files, new events appear, PEDF can easily handle it by extending the network, while structures such as RNN cannot handle such changes. One of the potential improvements of the model can be forgetting some less useful features. Adding a new event to the structure is straightforward, but removing a portion of the network requires more investigation.

Table 4, PEDF performance.

| Event error | | | | | | | | |
|---|---|---|---|---|---|---|---|---|
| Clusterer | Expected Maximization | | | | | | | |
| Known % | 20 | 30 | 40 | 50 | 60 | 70 | 80 | 90 |
| NB | 8,425 | 7,930 | 7,746 | 7,046 | 2,719 | 2,544 | 2,532 | 1,979 |
| RF | 6,022 | 5,192 | 4,696 | 2,597 | 1,881 | 1,775 | 1,770 | 402 |
| MLP | 5,858 | 5,348 | 4,933 | 2,717 | 1,502 | 1,371 | 1,361 | 584 |
| Clusterer | Canopy | | | | | | | |
| Known % | 20 | 30 | 40 | 50 | 60 | 70 | 80 | 90 |
| NB | 8,947 | 9,246 | 9,221 | 9,187 | 5,949 | 5,485 | 5,480 | 5,011 |
| RF | 8,314 | 7,946 | 7,826 | 7,480 | 2,969 | 2,807 | 2,796 | 2,296 |
| MLP | 8,914 | 8,914 | 8,823 | 8,793 | 3,181 | 2,960 | 2,945 | 2,546 |
| Clusterer | CascadeKMeans max 100 clusters | | | | | | | |
| Known % | 20 | 30 | 40 | 50 | 60 | 70 | 80 | 90 |
| NB | 9,274 | 8,969 | 8,799 | 8,314 | 3,246 | 3,148 | 3,135 | 2,002 |
| RF | 6,921 | 4,954 | 4,482 | 2,417 | 1,796 | 1,667 | 1,662 | 451 |
| MLP | 7,438 | 6,105 | 5,914 | 5,155 | 2,748 | 2,575 | 2,566 | 1,972 |
| Clusterer | Kmeans 50 clusters | | | | | | | |
| Known % | 20 | 30 | 40 | 50 | 60 | 70 | 80 | 90 |
| NB | 7,462 | 5,425 | 5,206 | 4,387 | 2,667 | 2,500 | 2,490 | 1,905 |
| RF | 6,945 | 5,086 | 4,658 | 1,922 | 1,194 | 984 | 977 | 306 |
| MLP | 6,912 | 3,992 | 3,838 | 2,979 | 2,612 | 2,434 | 2,423 | 1,768 |
| Duration error | | | | | | | | |
| Clusterer | Expected Maximization | | | | | | | |
| Known % | 20 | 30 | 40 | 50 | 60 | 70 | 80 | 90 |
| NB | 3,292,752 | 2,899,243 | 2,771,421 | 2,323,928 | 1,420,550 | 1,326,012 | 1,320,708 | 803,272 |
| RF | 3,047,081 | 2,157,690 | 2,122,273 | 1,763,782 | 1,397,120 | 1,020,134 | 1,015,869 | 108,530 |
| MLP | 2,128,846 | 1,857,891 | 1,821,554 | 1,247,516 | 1,083,049 | 783,874 | 779,403 | 200,318 |
| Clusterer | Canopy | | | | | | | |
| Known % | 20 | 30 | 40 | 50 | 60 | 70 | 80 | 90 |
| NB | 3,819,555 | 3,623,652 | 3,587,040 | 3,315,675 | 2,046,718 | 2,277,972 | 2,275,680 | 1,349,921 |
| RF | 3,742,058 | 3,268,028 | 3,235,717 | 2,987,298 | 1,502,247 | 1,818,813 | 1,812,006 | 940,493 |
| MLP | 3,827,364 | 3,721,745 | 3,681,598 | 3,400,189 | 1,616,739 | 1,961,669 | 1,952,068 | 1,030,496 |
| Clusterer | CascadeKMeans | | | | | | | |
| Known % | 20 | 30 | 40 | 50 | 60 | 70 | 80 | 90 |
| NB | 3,185,854 | 2,104,608 | 2,008,812 | 1,585,433 | 1,209,910 | 1,404,363 | 1,398,635 | 544,144 |
| RF | 2,569,700 | 2,147,744 | 2,151,174 | 1,666,581 | 1,337,491 | 1,160,463 | 1,154,833 | 144,300 |
| MLP | 2,998,078 | 2,631,354 | 2,562,767 | 2,212,105 | 1,376,950 | 1,156,288 | 1,152,019 | 433,850 |

| Clusterer | Kmeans 50 clusters | | | | | | | |
|---|---|---|---|---|---|---|---|---|
| Known % | 20 | 30 | 40 | 50 | 60 | 70 | 80 | 90 |
| NB | 2,676,887 | 1,974,686 | 1,901,820 | 1,533,806 | 1,293,236 | 1,049,323 | 1,045,136 | 542,569 |
| RF | 2,692,967 | 1,975,072 | 1,935,875 | 1,352,430 | 1,045,543 | 741,441 | 734,894 | 95,718 |
| MLP | 2,764,605 | 1,694,307 | 1,634,135 | 1,304,345 | 1,270,907 | 1,040,416 | 1,036,904 | 527,899 |
| Feature error | | | | | | | | |
| Clusterer | Expected Maximization | | | | | | | |
| Known % | 20 | 30 | 40 | 50 | 60 | 70 | 80 | 90 |
| NB | 484,340.14 | 336,530.82 | 310,157.15 | 322,558.92 | 243,897.60 | 229,836.59 | 228,601.63 | 48,466.63 |
| RF | 537,390.89 | 393,322.54 | 383,309.45 | 357,789.02 | 178,772.58 | 156,927.35 | 155,093.84 | 15,532.14 |
| MLP | 387,386.90 | 371,722.17 | 345,773.07 | 288,986.59 | 176,022.25 | 154,482.20 | 152,847.28 | 24,669.94 |
| Clusterer | Canopy | | | | | | | |
| Known % | 20 | 30 | 40 | 50 | 60 | 70 | 80 | 90 |
| NB | 436,297.98 | 447,869.52 | 445,751.14 | 440,712.44 | 327,623.03 | 284,551.60 | 283,304.03 | 207,007.84 |
| RF | 447,670.50 | 426,315.28 | 419,860.67 | 406,673.94 | 182,641.70 | 153,857.51 | 152,367.33 | 70,296.33 |
| MLP | 436,586.73 | 431,900.95 | 427,313.60 | 422,135.07 | 176,395.38 | 140,169.15 | 138,609.58 | 72,946.34 |
| Clusterer | CascadeKMeans | | | | | | | |
| Known % | 20 | 30 | 40 | 50 | 60 | 70 | 80 | 90 |
| NB | 480,140.11 | 440,707.44 | 415,334.01 | 396,439.10 | 226,112.48 | 205,383.30 | 204,316.52 | 53,425.72 |
| RF | 487,128.30 | 406,354.01 | 396,057.06 | 370,990.83 | 184,693.21 | 164,331.61 | 163,178.95 | 23,209.25 |
| MLP | 570,951.13 | 506,692.64 | 486,268.13 | 471,812.99 | 316,491.76 | 296,524.41 | 290,697.05 | 49,376.79 |
| Clusterer | Kmeans 50 clusters | | | | | | | |
| Known % | 20 | 30 | 40 | 50 | 60 | 70 | 80 | 90 |
| NB | 459,366.72 | 337,723.92 | 325,414.31 | 296,482.53 | 177,080.62 | 156,667.21 | 155,326.32 | 65,514.92 |
| RF | 440,862.10 | 341,878.37 | 310,407.88 | 211,657.10 | 138,264.56 | 116,420.32 | 115,346.53 | 14,305.36 |
| MLP | 446,884.22 | 271,041.91 | 262,902.02 | 230,518.45 | 163,521.83 | 140,330.30 | 138,904.41 | 39,854.28 |